\documentclass[conference]{IEEEtran}

\ifCLASSINFOpdf
\else
\fi
\usepackage{lipsum}
\usepackage{graphicx}
\usepackage{multirow}
\usepackage{dblfloatfix}

\usepackage[bookmarks=false]{hyperref}

\ifCLASSOPTIONcompsoc
\usepackage[caption=false,font=normalsize,labelfont=sf,textfont=sf]{subfig}
\else
\usepackage[caption=false,font=footnotesize]{subfig}
\fi

\usepackage{graphicx}
\usepackage{floatrow}
\usepackage{url}
\usepackage{textcomp}
\usepackage{amsmath}
\usepackage{multirow}
\usepackage{verbatim}
\usepackage{hyperref}
\usepackage[font=scriptsize]{caption}
\usepackage{subfig}

\newfloatcommand{capbtabbox}{table}[][\FBwidth]
\makeatletter
\def\endthebibliography{%
	\def\@noitemerr{\@latex@warning{Empty `thebibliography' environment}}%
	\endlist
}
\makeatother

\begin{document}
%
\title{A Robust Hybrid Approach for Textual Document Classification}



%

\author{\IEEEauthorblockN{Muhammad Nabeel Asim\IEEEauthorrefmark{1}\IEEEauthorrefmark{2}
Muhammad Usman Ghani Khan\IEEEauthorrefmark{2},\\
Muhammad Imran Malik\IEEEauthorrefmark{3}, 
Andreas Dengel\IEEEauthorrefmark{1},
Sheraz Ahmed \IEEEauthorrefmark{1}} 

Email: firstname.lastname@dfki.de

\IEEEauthorblockA{\IEEEauthorrefmark{1}German Research Center for Artificial Intelligence (DFKI), 67663 Kaiserslautern, Germany}
\IEEEauthorblockA{\IEEEauthorrefmark{2}National Center for Artificial Intelligence (NCAI), University of Engineering and Technology, Lahore,
Pakistan}
\IEEEauthorblockA{\IEEEauthorrefmark{3}National Center for Artificial Intelligence (NCAI), National University of Sciences and Technology,
Islamabad, Pakistan}}



\maketitle 



%
\IEEEpeerreviewmaketitle

\section{Abstract}

Text document classification is an important task for diverse natural language processing based applications. Traditional machine learning approaches mainly focused on reducing dimensionality of textual data to perform classification. This although improved the overall classification accuracy, the classifiers still faced sparsity problem due to lack of better data representation techniques. Deep learning based text document classification, on the other hand, benefitted greatly from the invention of word embeddings that have solved the sparsity problem and researchers’ focus mainly remained on the development of deep architectures. Deeper architectures, however, learn some redundant features that limit the performance of deep learning based solutions. In this paper, we propose a two stage text document classification methodology which combines traditional feature engineering with automatic feature engineering (using deep learning). The proposed methodology comprises a filter based feature selection (FSE) algorithm followed by a deep convolutional neural network. This methodology is evaluated on the two most commonly used public datasets, i.e., 20 Newsgroups data and BBC news data. Evaluation results reveal that the proposed methodology outperforms the state-of-the-art of both the (traditional) machine learning and deep learning based text document classification methodologies with a significant margin of 7.7\% on 20 Newsgroups and 6.6\% on BBC news datasets.

\begin{IEEEkeywords}
Text Document Classification, Filter based feature selection, 20 News Group, BBC News,   Multi-channel CNN 
\end{IEEEkeywords}

\section{Introduction}

Text classification is extensively being used in several applications such as information filtering, recommendation systems, sentiment analysis, opinion mining, and web searching \cite{aggarwal2012survey}. Broadly, text classification methodologies are divided into two classes statistical, and rule-based \cite{aziguli2017robust}. Statistical approaches utilize arithmetical knowledge, whereas rule-based approaches require extensive domain knowledge to develop rules on the basis of which samples could be classified into a predefined set of categories. Rule-based approaches are not extensively being used because it is a difficult job to develop robust rules which do not need to update periodically.  

Previously, researchers performed automatic document text classification by using machine learning classifiers such as Naive Bayes \cite{langley1992analysis}, SVM, NN, Decision Trees \cite{peterson2009k, luo2017quantized}. In recent years, a number of feature selection algorithms have been proposed which significantly improve the performance of text classification \cite{rehman2017feature,guyon2003introduction,lal2006embedded, wald2013filter}. Although feature selection techniques reduce the dimensionality of data to a certain level, however still traditional machine learning based text classification methodologies face the problem of feature representation as trivial feature representation algorithms use a bag of words model which consider unigrams, n-grams or specific patterns as features \cite{le2014distributed}. Thus, these algorithms do not capture the complete contextual  information of data and face the problem of data sparsity. 

 The problem of data sparsity is solved by word embeddings which do not only capture syntactic but semantic information of textual data as well \cite{mikolov2013distributed}. Deep learning based text classification methodologies are not only successfully capturing the contextual information of data, but also resolving the data sparsity problems, thus, they are outperforming state-of-the-art machine learning based classification approaches \cite{jiang2018text,shih2017investigating}.

Primarily in computer vision and NLP, researchers have been trying to develop deeper neural network architectures which could extract a better set of features for classification \cite{szegedy2015going, lee2015deeply}. However, deeper architectures are not only computationally more expensive but complicated relationships learned by deeper architectures will actually be the outcome of sampling noise in case of small scale datasets. Recent researches showed that deeper architectures extract redundant features which eventually reduce the classification performance \cite{denil2013predicting, ayinde2016clustering, ayinde2019correlation}.

 This paper proposes a two stage text classification(TSCNN) methodology, which is a hybrid approach. The first stage relies on the feature selection algorithm where the aim is to rank and remove all irrelevant and redundant features. While the second stage is based on deep learning, where  from first stage discriminative features are fed to multi-channel CNN model. In this novel setting, the proposed approach reap the benefits of both traditional feature engineering and automated feature engineering (using deep learning).
 Extensive evaluation of two commonly used publicly available datasets reveals that the proposed approach outperforms state-of-the-art methods with a significant margin.

\section{Related Work}\label{rlwork}
This section provides a birds-eye view on state-of-the-art filter based feature selection algorithms used in statistical based text document classification approaches. Moreover, recent deep learning based text classification methodologies are also briefly described.

Feature selection is considered an indispensable task in text classification as it removes redundant and irrelevant features of the corpus \cite{asim2017comparison}. Broadly, feature selection approaches can be divided into three classes namely wrapper, embedded, and filter \cite{lal2006embedded, wald2013filter}. In recent years, researchers have proposed various filter based feature selection methods to raise the performance of document text classification \cite{rehman2018selection}.

 Document frequency \cite{dasgupta2007feature} is the simplest metric used to rank the features in training data by utilizing the presence of a certain feature in positive and negative class documents respectively. Another simplest feature selection algorithm namely Accuracy ($ACC$) is the difference between true positive and false positive of a feature \cite{forman2003extensive}. $ACC$ is biased towards true positive ($t_p$) because it assigns a higher score to those features which are more frequent in positive class. In order to tackle the biaseness, an advanced version of $ACC$ namely Balanced Accuracy Measure (ACC2) was introduced which is based on true positive rate ($t_{pr}$) and false positive rate ($f_{pr}$). Although $ACC2$ resolves the issue of class unbalance through normalizing true and false positives with the respective size of the classes, however, $ACC2$ assigns the same rank to those features which reveal same difference value ($|t_{pr}-fpr|$) despite having different $t_{pr}$ or $f_{pr}$.

Furthermore, Information Gain ($IG$) is another commonly used feature selection algorithm in text classification \cite{xu2008efficient}. It determines whether the information required to predict the target class of a document is raised or declined through the addition or elimination of a feature. Likewise, Chi-squared (CHISQ) considers the existence or non-existence of a feature to be independent of class labels. CHISQ does not reveal promising performance when the dataset is enriched with infrequent features, however, its results can be raised through pruning \cite{forman2003extensive, asim2017effect}.

Odds Ratio (OR) \cite{bland2000odds} is the likelihood ratio among the occurrence of a feature and the absence of a feature in the certain document. It gives the highest rank to the rare features. Thus, it performs well with fewer features, however, its performance starts getting deteriorated with the increasing number of features. Similarly,Distinguish Feature Selector (DFS) \cite{uysal2012novel} considers those features to be more significant which occur more frequently in one class and less frequent in other classes. Furthermore, Gini Index is used to estimate the distribution of a feature over given classes. Although it was originally used to estimate the GDP\_per\_Capita, however, in text classification, it is used to rank the features \cite{singh2010new}. 

It is considered that deep learning models automate the process of feature engineering, contrarily, recent research in computer vision reveals that deep learning models extract some irrelevant and redundant features \cite{ayinde2019correlation}. In order to raise the performance of text document classification, several researchers have utilized diverse deep learning based methodologies. For instance, Lai et al. \cite{lai2015recurrent} proposed a Bi-directional recurrent structure in a convolutional neural network for text classification. This recurrent structure captures the contextual information while learning word representations and produced less noise as compared to the trivial window based convolutional network. Moreover, a max pooling layer was used in order to select highly significant words. Through combining recurrent structure and max-pooling layer, they utilized the benefits of both convolutional and recurrent neural networks. The approach was evaluated on sentiment analysis, topic classification, and writing style classification.


Aziguli et al. \cite{aziguli2017robust} utilized hybrid deep learning methods and proposed denoising deep neural network (DDNN) based on restricted Boltzmann machine (RBM), and denoising autoencoder (DAE). DDNN alleviated noise and raised the performance of feature extraction. Likewise, in order to resolve the problem of computing high dimensional sparse matrix for the task of text classification, Jiang et al. \cite{jiang2018text} proposed hybrid text classification model which was utilizing deep belief network (DBN) for feature extraction, and softmax regression to classify given text. They claimed that the proposed hybrid methodology performed better than trivial classification methods on two benchmark datasets.

\begin{figure*}[!h]
	\centering
	\includegraphics[width=0.82\textwidth]{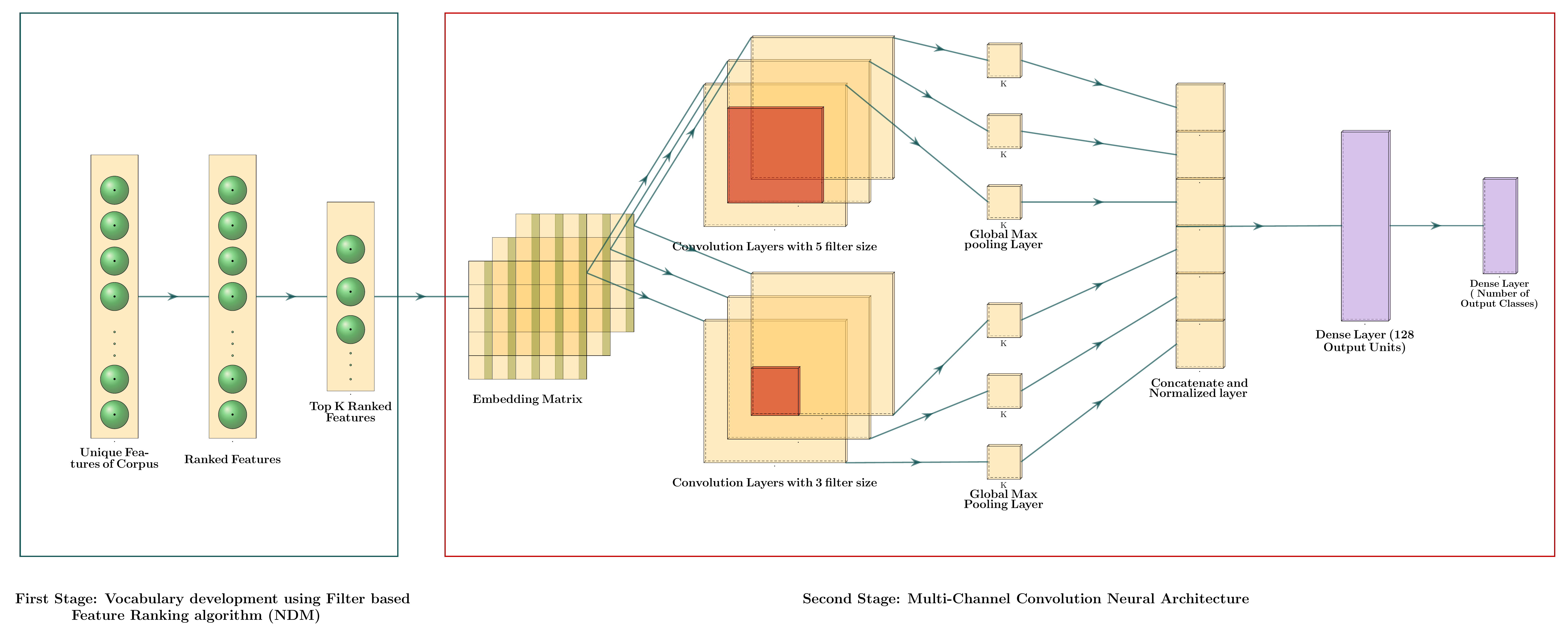}
	\caption{Proposed Two Stage Classification Methodology}
	\label{conv}
\end{figure*}

Moreover, Huang et al. \cite{huang2014research} utilized deep belief networks in order to acquire emotional features from speech signals. Extracted features were fed to non-linear support vector machine (SVM) classifier and in this way a hybrid system was established for the task of identifying emotions from speech. Zhou et al. \cite{zhou2014active} presented an algorithm namely active hybrid deep belief network (semi-supervised) for the task of sentiment classification. In their two fold network, first, they extracted features using restricted Boltzmann machines and then preceding hidden layers learned the comments using convolutional RBM (CRBM).

 Kahou et al. \cite{kahou2013combining} revealed that dropout performance could be further enhanced by using Relu unites rather than max-out units. Srivastava et al \cite{srivastava2014dropout} revealed that dropout technique raises the performance of all neural networks on several supervised tasks like document classification, speech recognition, and computational biology.

Liu et al. \cite{liu2017leveraging} presented an attentional framework based on deep linguistics. This framework incorporated concept information of corpus words into neural network based classification models. MetaMap and WordNet were used to annotate biomedical and general text respectively.  Shih et al. \cite{srivastava2014dropout} purposed the novel use of Siamese long short-term memory (LSTM) based deep learning method to better learn document representation for the task of text classification.

\section{Methodology}\label{methdology}

This section briefly describes the proposed methodology of two stage text classification shown in Figure \ref{conv}. First stage, is dedicated for feature selection where irrelevant and redundant features are removed using Normalized Difference Measure ($NDM$). While second stage use multi-channel CNN model for the classification of textual documents into predefined categories based on discriminative patterns extracted by convolution layers.

\subsection{Discriminative Feature Selection}

To develop the vocabulary of most discriminative features, we remove all punctuation symbols and non-significant words (stop words) as a part of the preprocessing step. Furthermore, in order to rank the terms based on their discriminative power among the classes, we use filter based feature selection method named as Normalized Difference Measure (NDM)\cite{rehman2017feature}. Considering the features contour plot, Rehman et al. \cite{rehman2017feature} suggested that all those features which exist in top left, and bottom right corners of the contour are extremely significant as compared to those features which exist around diagonals. State-of-the-art filter based feature selection algorithms such as ACC2 treat all those features in the same fashion which exist around the diagonals \cite{rehman2017feature}. For instance, ACC2 assigns same rank to those features which has equal difference ($|t_{pr} -f_{pr}|$) value but different $t_{pr}$ and $f_{pr}$ values. Whereas NDM normalizes the difference ($|t_{pr} -f_{pr}|$) with the minimum of $t_{pr}$ and $f_{pr}$ (min($t_{pr}$, $f_{pr}$)) and assign different rank to those terms which have same difference value. Normalized Difference Measure (NDM) considers those features highly significant which have the following properties:

\begin{itemize}
	\item High $|t_{pr} - f_{pr}|$ value.
	\item $t_{pr}$ or $f_{pr}$ must be close to zero.
	\item If two features got the same difference  $|t_{pr} - t_{pr}|$ value, then a greater rank shall be assigned to that feature which reveal least min($t_{pr}$, $f_{pr}$) value. 
	
\end{itemize}

Mathematically NDM is represented as follows:

\begin{equation}
NDM =\frac{|t_{pr}-f_{pr}|}{min(t_{pr},f_{pr})}
\end{equation}

where $t_{pr}$ refers to true positive rate and $f_{pr}$ refers to false positive rate.
True positive rate is the ratio between the number of positive class documents having term t and the size of positive class. False positive rate is the ratio between the number of negative class documents having term t and the size of negative class.

\subsection{Multi-Channel CNN Model}

In second stage, a convolutional neural network (CNN) based on three channel is used. Each channel has two wide convolutional layers with 16 filters of size 5 and 3 respectively. We use multi-Channel CNN model to extract a variety of features at each channel by feeding different representation of features at the embedding layer. The first channel contains features obtained from  FastText embedding provided by Mikolov et al. \cite{mikolov2018advances}. These pre-trained word vectors were developed after training the skip-gram model on Wikipedia 2017 documents, UMBC web base corpus, and statmt.org news dataset using Fasttext\footnote{https://fasttext.cc/} API. There are total one million words provided with pre-trained word vectors of dimension 300, whereas the other two channels are exploiting randomly initialized embedding layers. Finally, the features of all three channels are concatenated to form a single vector.  All wide convolution layers are using $Tanh$ as activation function and allow every feature to equally take part while convolving. Each convolution layer is followed by a global max pooling layer which extracts the most discriminative feature from yielded feature maps. After global max pooling, all discriminative features are concatenated and normalized using L2 normalization technique. These normalized features are then passed to a fully connected layer which has 128 output units and using relu as the activation function. Finally, last fully connected layer use softmax as activation function and acts as a classifier.

\section{Experimental Setup}\label{exp}

This section describes the experimental setup used to evaluate the integrity of proposed text classification methodology on two benchmark datasets namely BBC News and 20 NewsGroup.

In our experimentation, CNN is trained on two different versions of each dataset. In the first version named as Standard CNN (SCNN), the entire vocabulary of each dataset obtained after preprocessing is fed to the model. Whereas, in the second version named as Two Stage CNN (TSCNN), after preprocessing, the vocabulary of each class is ranked using filter based feature selection algorithm namely NDM and then only top k ranked features of each class are selected to feed the embedding layer of underlay model. Top 1000 features of BBC, and 10,000 features of 20 Newsgroup dataset are selected and only these selected features are fed to the embedding layer of each channel. Furthermore, as 20 Newsgroup dataset has more unique features as compared to BBC dataset, so in final vocabulary, for 2o Newsgroup dataset we select more features as compared to BBC news dataset.  Keeping only top 1000, and 10,000 features, two vocabularies of size 4208 and 41701 are constructed for respective datasets. Since the features are ranked on class level, therefore, many features overlap in different classes.

For experiments, we use 20 newsgroup dataset which has a standard split of 70\% training samples, and 30\% test samples. We use 10\% of training samples for validation. Moreover, BBC news dataset has no standard split, therefore, we consider 60\% of data for training, 10\% for validation, and 30\% for testing.

Table \ref{stats} summarizes the statistics of two datasets (20NewsGroup, BBC News) used in our experimentation.

\begin{table}[!h]
	\centering
	\resizebox{1.0\textwidth}{!}	
	{\begin{tabular}{|c|c|c|c|c|c|}
			\hline
			\textbf{Dataset} & \textbf{Total Documents} & \textbf{Number of features} & \textbf{Number of Classes} & \textbf{Min Class Size} & \textbf{Max Class Size}                                                                                                 \\ \hline
			20 NewsGroup     & 18846                    & 41520                       & 20                         & 628                     & 998                      \\ \hline
			BBC News             & 2225                     & 34318                       & 5                          & 386                     & 511                                                                                                      \\ \hline
	\end{tabular}}
	\caption{Statistics of 20newsgroup and BBC datasets}
	\label{stats}
\end{table}

 RMSprop is used as an optimizer with learning rate of 0.001 and categorical cross-entropy is used as a loss function. Batch size of 50 is used and we train the model for 20 epochs.

\section{Results}\label{rel}

This section provides detailed insight and analysis of several experiments performed to uncover pros and cons of the proposed approach in comparison standard CNN model (SCNN). To evaluate the effect of irrelevant and redundant features on the performance of convolutional neural network, we have also shown confusion matrices to reveal the performance of two stage classification and standard CNN methodologies. Moreover, we also compare the performance of proposed two stage classification methodology with the state-of-the-art machine and deep learning based text classification methodologies.

\begin{figure}[!h]
	\centering
	\includegraphics[width=0.9\textwidth]{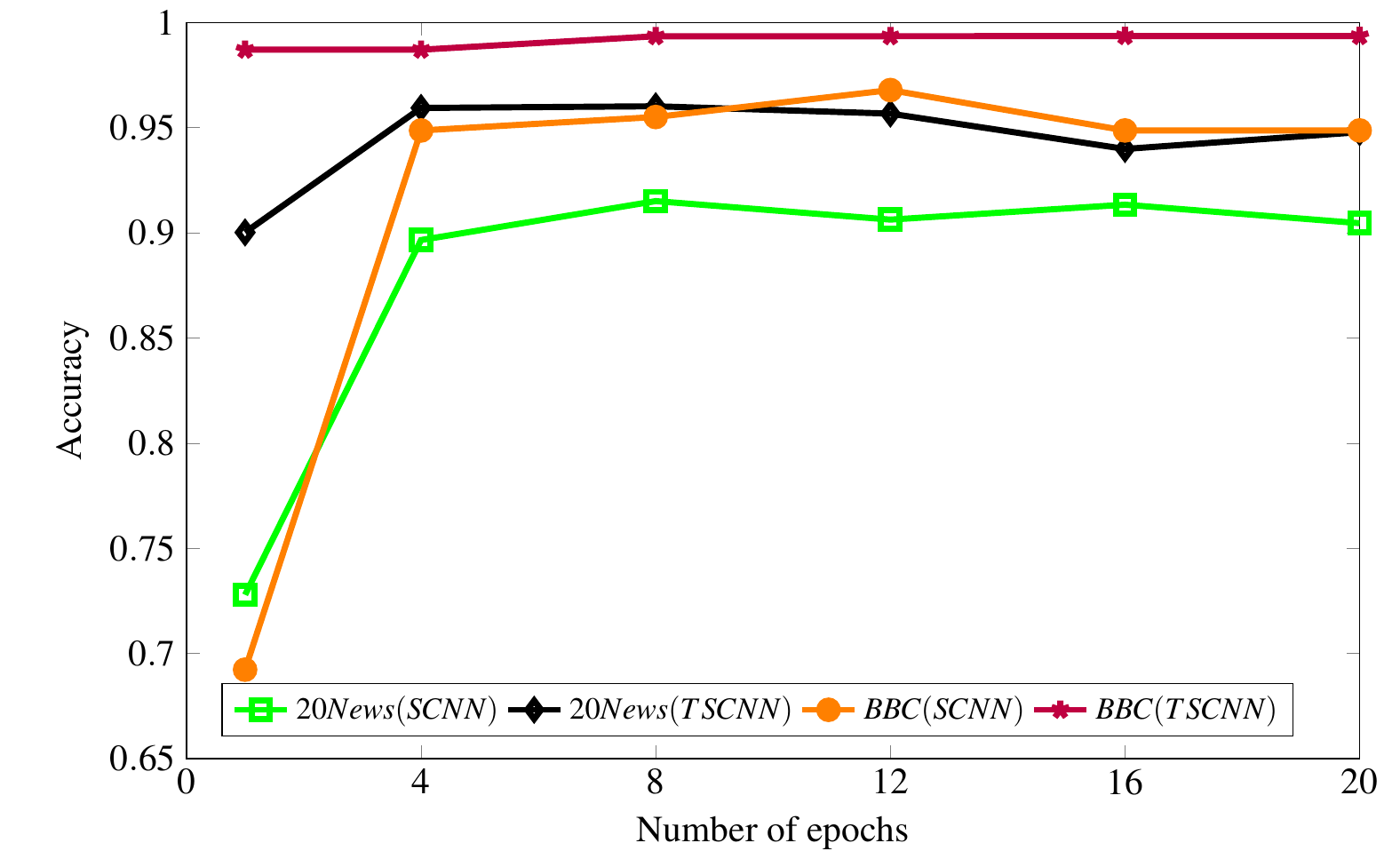}
	\caption{Accuracy values produced by  two stage classification methodology, and standard CNN model on validation set of two benchmark datasets 20 newsgroup, and BBC news}
   \label{fig:20newsacc1}

\end{figure}

Figure \ref{fig:20newsacc1} shows the accuracy of proposed two stage classification, and standard CNN classification methodologies on the validation set of 20 newsgroup, and BBC news datasets respectively.

For 20 newsgroup dataset, the accuracy of standard CNN classification methodology begins at low of 73\% as compared to the accuracy of two stage (TSCNN) classification methodology which reveals a promising figure of 90\%. This performance gap occurs due to lack of discriminative features in standard CNN. TSCNN is fed with highly discriminative features, whereas the standard CNN model extracts significant features from given vocabulary on its own. This is why standard CNN performance gets improve until 4 epochs as compared to TSCNN whose performance increases slightly. However, the standard CNN model still does not manage to surpass the promising performance of TSCNN at any epoch.

Likewise, for BBC news dataset, both models depict similar performance trend as discussed for 20 newsgroup dataset.

\begin{figure}[!tph]

	\centering
     \includegraphics[width=0.9\textwidth]{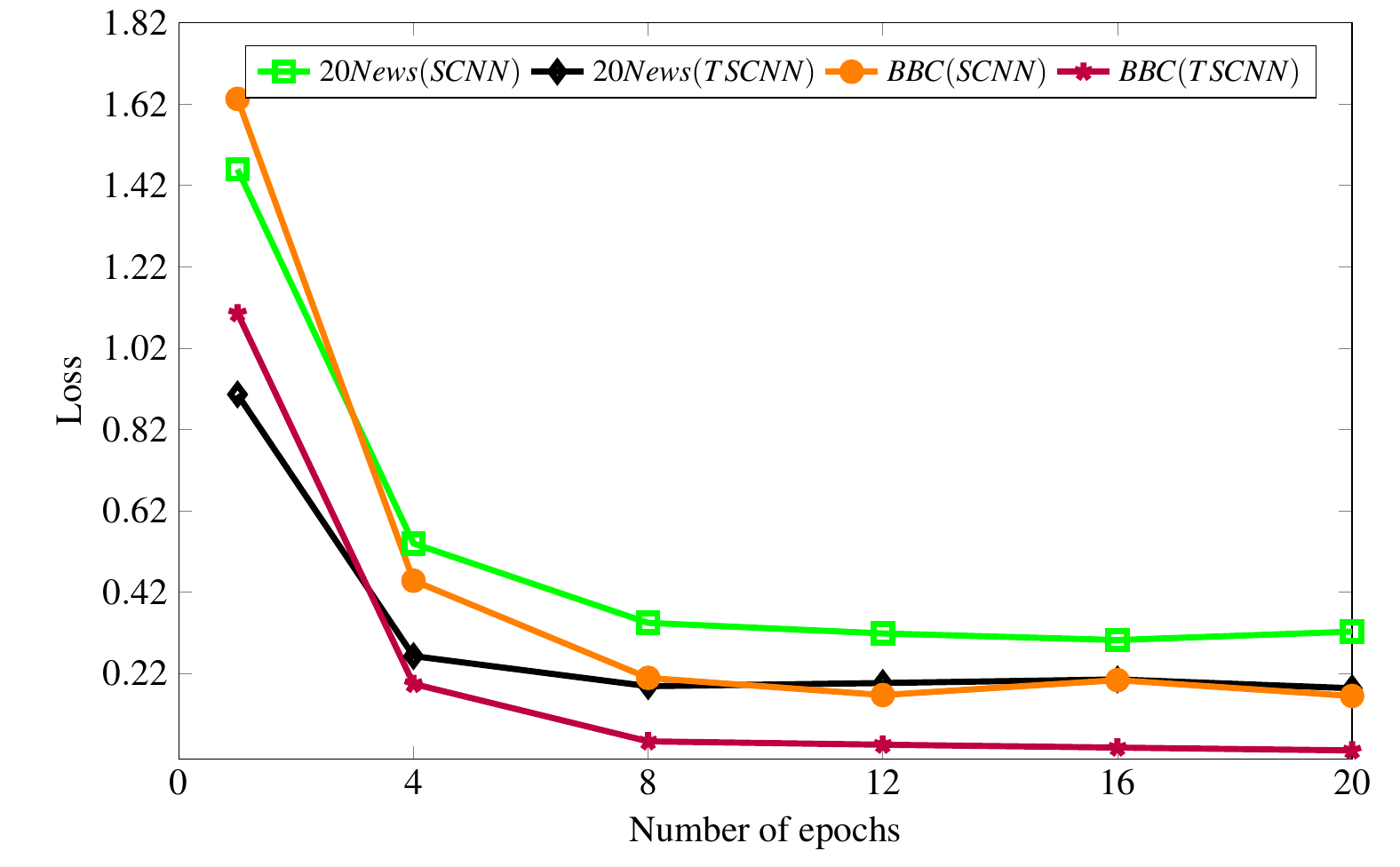}
	\caption{Loss values produced by  two stage classification methodology, and standard CNN model on validation set of two benchmark datasets 20 newsgroup, and BBC news}
	\label{fig:20newsacc22}
	
\end{figure}

Figure \ref{fig:20newsacc22} compares the loss values produce by TSCNN and SCNN at different epochs of two datasets.

For 20 newsgroup dataset, at first epoch, there is a difference of 0.55 between the loss values of TSCNN and SCNN, due to the fact that the vocabulary of unique words fed to TSCNN is noise free and it has to learn more discriminative features from a vocabulary of irrelevant and redundant features. On the other hand, complete vocabulary was fed to SCNN which contains both relevant and irrelevant features.
The assumption was that SCNN will automatically select relevant features and discard which are unimportant. Furthermore, SCNN was unable to remove noise effectively since there is a gap of almost 0.2 between the losses of SCNN and TSCNN after 8th epochs.

Similarly, for BBC news dataset, both models have revealed a similar trend as discussed for 20 newsgroup dataset.

To evaluate the effect of noise on the performance of TSCNN and SCNN we have shown confusion matrices for both datasets. 

\begin{figure*}[!tph]
	\minipage{0.5\textwidth}
	\centering
	\includegraphics[width=1\linewidth]{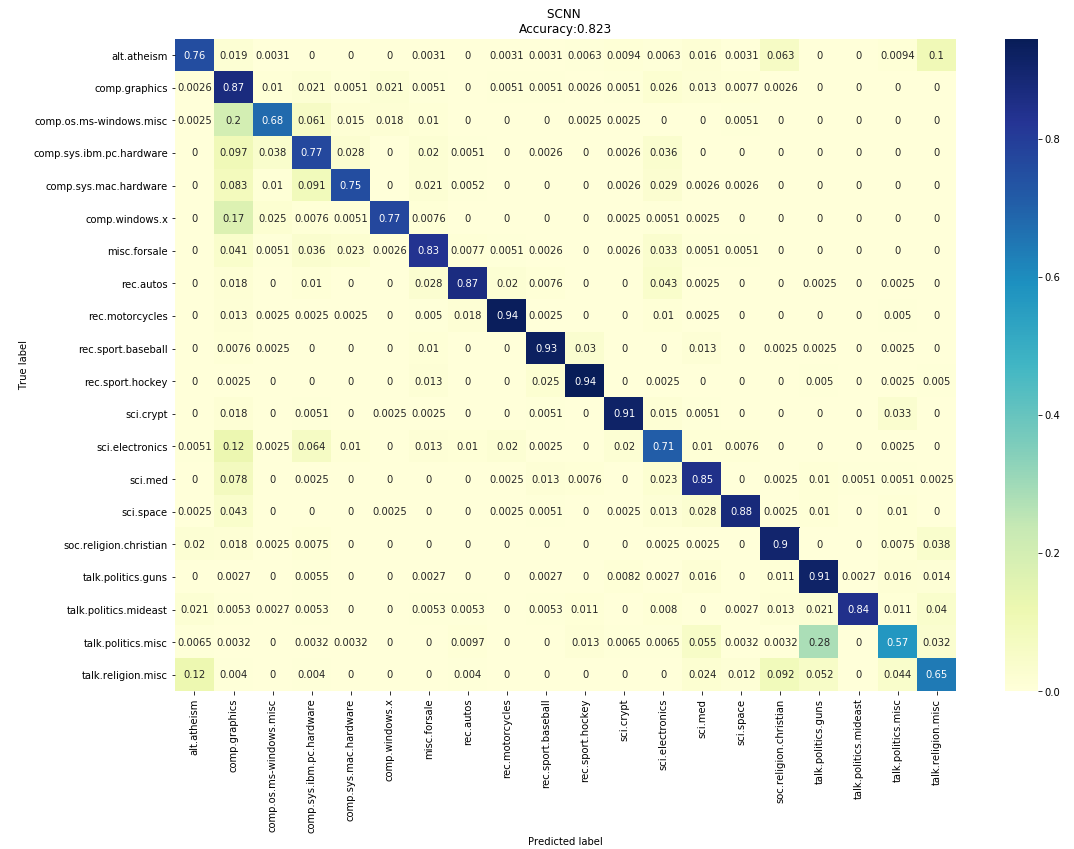}
	\caption{Confusion Matrices of TSCNN and SCNN for 20 Newsgroup dataset }
	\label{CMBBC}
	\endminipage\hfill
	\minipage{0.5\textwidth}
	\centering
	\includegraphics[width=1\linewidth]{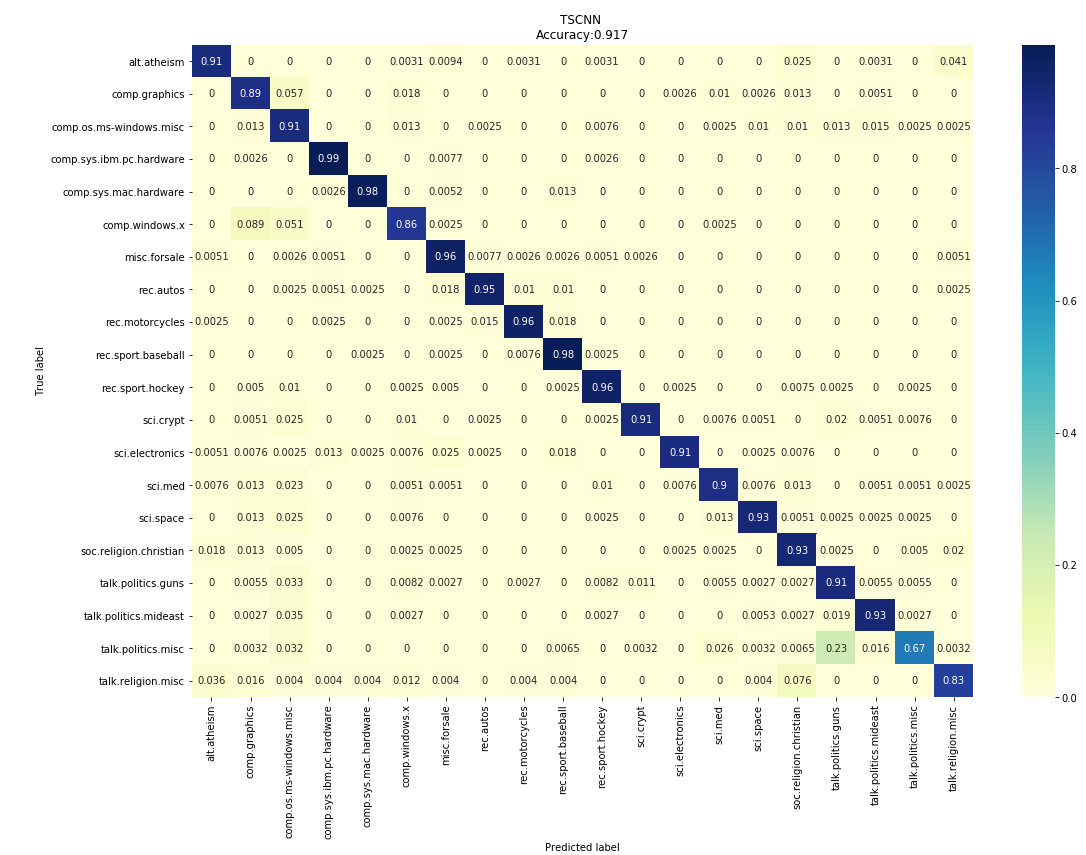}
	\endminipage\hfill
\end{figure*}

Figure \ref{CMBBC} illustrates that in the case of standard CNN classes like  \textit{talk.politics.misc} and \textit{talk.religion.misc} are slightly confused with classes \textit{talk.politics.guns} and \textit{alt.atheism} respectively. However classes like \textit{sci.electronics} and \textit{comp.graphics} are confused with many other classes. 


On the other hand, confusion matrix of TSCNN confirms that the confusion between the classes is resolved by using two stage classification methodology which develops a vocabulary of discriminative words. It can also be confirmed by observing the accuracy of the classes such as \textit{talk.religion.misc}, \textit{sci.electronics} and \textit{comp.os.ms-windows.misc} were increased from 61\%, 69\% and 74\% to 83\%, 91\% and 92\% respectively. 

\begin{figure}[!tph]
	\minipage{0.5\textwidth}
	\centering
	\includegraphics[width=1\linewidth]{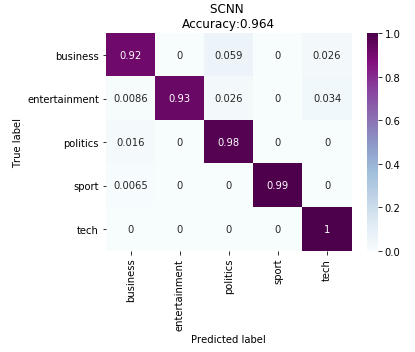}
	\caption{Confusion Matrices of TSCNN and SCNN for BBC News dataset }
	\label{CM22}
	\endminipage\hfill
	\minipage{0.5\textwidth}
	\centering
	\includegraphics[width=1\linewidth]{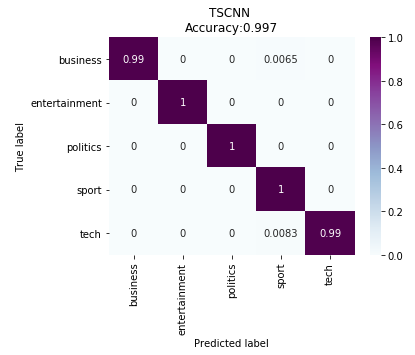}
	\endminipage\hfill
\end{figure}

Similarly, the confusion matrices for BBC News Datasets are also shown in figure \ref{CM22}, which also demonstrate the same phenomenon mentioned before. As it can be clearly seen that \textit{business} and \textit{entertainment} classes are confused  with other classes when classified using standard CNN. Whereas, two stage classification discard the inter-class dependencies almost completely as the accuracy of \textit{business} and \textit{entertainment} classes increases from 92\% to 99\% and 100\% respectively.



\subsection{ Comparison with state-of-the-art}

This section provides insight into the presented hybrid approach in comparison to the state-of-the-art machine and deep learning based methodologies.

\begin{table*}[t]
	\centering
	\resizebox{1.0\textwidth}{!}{
		\begin{tabular}{|l|l|l|l|l|l|}
			\hline
			\multirow{3}{*}{Text Classification System}                 &   \multirow{3}{*}{Method}  & \multicolumn{4}{c|}{Dataset}                                                                                                                                      \\ \cline{3-6} 
			&  & \multicolumn{2}{c|}{BBC News} & \multicolumn{2}{c|}{20 News group}                                                                                                \\ \cline{3-6} 
			& &  Accuracy (\%)   & F1 Measure  & Accuracy                      & F1 Measure                                                                                        \\ \hline
			Rehman et al. [2017] \cite{rehman2017feature}  & \begin{tabular}[c]{@{}l@{}}NDM, \\ SVM, NB     \end{tabular}             & -             & -         & -                           & 75.1                                                                                              \\ \hline
			Rehman et al. [2018] \cite{rehman2018selection}  & \begin{tabular}[c]{@{}l@{}}MMR, \\ SVM, NB     \end{tabular}                   & -             & -         & -                           & 84.0                                                                                              \\ \hline
			Lai et al. [2015b] \cite{lai2015recurrent}  & Modified CNN              & -             & -         & -                           & \begin{tabular}[c]{@{}l@{}}96.49 on 4 classes \\ (comp, politics, rec, and religion)\end{tabular} \\ \hline
			Aziguli et al. [2017b]\cite{aziguli2017robust}     & \begin{tabular}[c]{@{}l@{}}Auto encoder \end{tabular}             & 92.86           & 92.67       & 73.78                         & 73.49                                                                                             \\ \hline
			Jiang et al. [2018b]\cite{jiang2018text}       &DBN+Softmax          & -             & -         & 85.57                         & -                                                                                               \\ \hline
			Liu et al. [2017b]\cite{liu2017leveraging}    & \begin{tabular}[c]{@{}l@{}} Deep Learning and \\ meta-thesaurus \end{tabular}                     & -             & -         & 69.82                         & -                                                                                               \\ \hline
			Shih et al. [2017b]\cite{shih2017investigating}    &LSTM             & -             & -         & 86.2                          & -                                                                                               \\ \hline
			Shirsat et al. [2019]\cite{shirsat2019sentence}        & SVM, NB              & 96.46           & -         & -                           & -                                                                                               \\ \hline
			Camacho [2017] \cite{camacho2017role}         & CNN, LSTM              & 97.0            & -         & 90.9 for topic categorization & -                                                                                               \\ \hline
			Pradhan et al. [2017] \cite{pradhan2017comparison}    &ML classifiers Comparison             & 97.67           & -         & 86.70                         & -                                                                                               \\ \hline
			Elghannam [2019] \cite{elghannam2019text}     &Feature representation, ML classifier                 & 92.6            & 92.6        &                               & -                                                                                               \\ \hline
			Wang et al. [2019] \cite{wang2019softly}         & Cross domain Transfer learning             & -             & -         & 95.62 (six categories)        & -                                                                                               \\ \hline
			Standard CNN Model &  Multi Channel CNN & 94.6            & 94.4        & 82.76                         & 82.57                                                                                             \\ \hline
			\begin{tabular}[c]{@{}l@{}}Proposed Two Stage \\ Classification Methodology \end{tabular} & Feature Engineering, Multi Channel CNN & \textbf{99.251}          & \textbf{99.256}      & \textbf{91.729}                        & \textbf{91.746}                                                                                            \\ \hline
	\end{tabular}}
	\caption {Performance Comparison of two stage classification methodology  with state-of-the-art machine and deep learning methodologies on two bench mark datasets in terms of Accuracy and $F_1$ measure }
	\label{tablecomp}
\end{table*}

Table \ref{tablecomp} illustrates the results of the proposed methodology, and 12 well known methods from literature including state-of-the-art results produced by machine and deep learning methodologies on 20 newsgroup and BBC news datasets.

In order to improve the performance of machine learning based text classification, Rehman et al. \cite{rehman2017feature} proposed a filter based feature selection algorithm namely Normalized Difference Measure (NDM). They compared its performance with seven state-of-the-art feature selection algorithms (ODDS, CHI, IG, DFS, GINI, ACC2, POISON) using SVM, and Naive Bayes classifiers. Their experimentation proved that the removal of irrelevant and redundant features improves the performance of text classification. They reported the highest macro $F_1$ score of 75\% on 20 newsgroup dataset. Lately, Rehman et al. \cite{rehman2018selection} proposed a new version of NDM and named it as MMR. MMR outperformed NDM with the figure of 9\%. Moreover, Shirsat et al. \cite{shirsat2019sentence} performed
sentiment identification on sentence level using positive, and negative words list provided by Bing Liu dictionary. Their proposed methodology marked the performance of 96\% with SVM classifier on BBC news dataset. Recently, Pradhan et al. \cite{pradhan2017comparison} compared the performance of several classification algorithms (SVM, Naive Bayes, Decision Tree, KNN, Rocchio) on number of news datasets. They extrapolated that SVM outperformed other four classifiers on all datasets. SVM produced the performance figure of 86\%, and 97\% on 20 newsgroup and BBC news datasets. Elghannam \cite{elghannam2019text} used the bi-gram frequency for the representation of the document in a typical machine learning based methodology. The proposed approach did not require any NLP tools and alleviated data sparsity up to great extent. They reported the $f_1$ score of 92\% on BBC news dataset.

Wang et al. \cite{wang2019softly} presented transfer learning method in order to perform text classification for cross-domain text. They performed experimentation on six classes of 20 newsgroup dataset and managed to produce the performance of 95\%.

On the other hand, researchers have utilized deep learning based diverse methodologies to raise the performance of text classification. For instance, the convolutional neural network based on Bi-directional recurrent structure \cite{lai2015recurrent} successfully extracted the semantics of underlay data. It produced the performance of 96.49\% on four classes (politics,comp,religion,rec) of 20 newsgroup dataset. Likewise, Aziguli et al. \cite{aziguli2017robust} proposed denoising deep neural networks exploited restricted boltzmann machine and denoising autoencoder to produce the performance of 75\%, and 97\% on 20 newsgroup, and BBC datasets respectively. Whereas, deep belief network and softmax regression were combinely used \cite{jiang2018text} to select discriminative features for text classification. Combination of both managed to mark the accuracy of 85\% on 20 newsgroup dataset. Moreover, a deep linguistics based framework \cite{liu2017leveraging} utilized WordNet, and MetaMap to extend concept information of underlay text. This approach produced the accuracy of just 69\% on 20 newsgroup dataset. Similarly, in order to improve the learning of document representation, siamese long short-term memory (LSTM) based deep learning methodology was proposed \cite{shih2017investigating} which revealed the performance of 86\% on 20 newsgroup dataset. Camacho-Collados and Pilehvar\cite{camacho2017role} revealed effective preprocessing practices in order to train word embeddings for the task of topic categorization. Their experimentation utilized two versions of CNN namely standard CNN with ReLU, and standard CNN with the addition of recurrent layer (LSTM) to produce the accuracy of 97\% on  BBC, and 90\% on 20 newsgroup datasets using 6 classes only.

The proposed two stage classification methodology has outperformed state-of-the-art machine and deep learning based methodologies. Moreover, in order to reveal the impact of feeding discriminative feature based vocabulary, we compare the proposed two stage classification methodology with the standard convolutional neural network.  

 Table \ref{tablecomp} clearly depicts that simple CNN produces the $F_1$ score of  94\%, and 82\% on BBC, and 20 newsgroup datasets respectively using SCNN. Whereas, TSCNN reveals the f1-score of 99\% and 91\% on BBC, and 20 newsgroup datasets using the set of features ranked by NDM.

\section{Conclusion}\label{cun}
This paper proposes a two stage classification methodology for text classification.  Firstly, we employ filter based feature selection algorithm(NDM) to develop noiseless vocabulary. Secondly, this vocabulary is fed to a multi-channel convolutional neural network, where each channel has two filters of size 5, and 3 respectively and 2 dense layers. Trivial convolutional layers, do not convolve all the features equally, this is why wide convolutional layers are used. Experimental results reveal that instead of feeding the whole vocabulary to CNN model, vocabulary of most discriminative features produces better performance.  
In future, we will assess the performance of the proposed two stage classification methodology using RNN, and Hybrid deep learning methodologies. Moreover, other renowned feature selection algorithms will be applied in first stage of proposed methodology.




\bibliographystyle{IEEEtran}

\bibliography{IEEEabrv,ram}
%

\end{document}